\pdfoutput=1
\documentclass[11pt]{article}

\usepackage{EMNLP2022}

\usepackage{times}
\usepackage{latexsym}
\usepackage{color, colortbl}
\usepackage[T1]{fontenc}
\usepackage{subcaption}
\usepackage[utf8]{inputenc}
\usepackage{microtype}
\usepackage{todonotes}
\usepackage{multirow}
\usepackage{xurl}

\definecolor{Gray}{gray}{0.9}

\title{Cross-lingual neural fuzzy matching for exploiting\\ target-language monolingual corpora in computer-aided translation}

\author{Miquel Esplà-Gomis, Víctor M. Sánchez-Cartagena\\
 {\bf Juan Antonio Pérez-Ortiz}, {\bf Felipe Sánchez-Martínez} \\[1ex]
Dep. de Llenguatges i Sistemes Informàtics \\
Universitat d'Alacant\\
E-03690 Sant Vicent del Raspeig (Spain)\\
{\tt \{mespla,vmsanchez,japerez,fsanchez\}@dlsi.ua.es}}

\begin{document}
\maketitle

\begin{abstract}  
Computer-aided translation (CAT) tools based on translation memories (MT) play a prominent role in the translation workflow of professional translators. However, the reduced availability of in-domain TMs, as compared to in-domain monolingual corpora, limits its adoption for a number of translation tasks. In this paper, we introduce a novel neural approach aimed at overcoming this limitation by exploiting not only TMs, but also in-domain target-language (TL) monolingual corpora, and still enabling a similar functionality to that offered by conventional TM-based CAT tools. Our approach relies on cross-lingual sentence embeddings to retrieve  translation proposals from TL monolingual corpora, and on a neural model to estimate their post-editing effort. The paper presents an automatic evaluation of these techniques on four language pairs that shows that our approach can successfully exploit monolingual texts in a TM-based CAT environment, increasing the amount of \emph{useful} translation proposals, and that our neural model for estimating the post-editing effort enables the combination of translation proposals obtained from monolingual corpora and from TMs in the usual way. A human evaluation performed on a single language pair confirms the results of the automatic evaluation  and  seems to indicate that the translation proposals retrieved with our approach are more useful than what the automatic evaluation shows.
\end{abstract}

\section{Introduction} \label{s:intro}
Despite the recent advances in machine translation (MT), \emph{translation memories} (TM) play a prominent role in the translation workflows of language service providers, translation departments, and freelance translators; the 2020 edition of the annual European Language Industry Survey~\citep{elis20} reported TMs to be the most popular resource among freelance translators and a relevant target of investment for companies and translation departments. Even when the perceived quality of the translation proposals from TM and MT is quite similar, professional translators need more time to post-edit MT outputs than TM proposals \cite{pilarandy2019}. In addition, the use of TMs foster terminology consistency and is especially suited to repetitive translation tasks as they allow for convenient recycling of previous translations.

Formally, a TM~\cite{isabelle1993} may be defined as a collection of translation units (TU), that is, source-target sentence pairs $(s,t)$ which are mutual translations. They are conveniently exploited by \emph{computer-aided translation} (CAT) tools which, for each new sentence to be translated $s'$, retrieve from the TM the subset of TUs $\{(s,t)\}$ whose source sentences $s$ are most \emph{similar} to $s'$. Translators then choose the TU that better fits their needs and make the appropriate changes to $t$, the target sentence in the selected TU, to convert it into $t'$, an adequate translation of $s'$.  

The similarity between 
$s'$ and the source sentence $s$ in a TU is estimated by means of the \emph{fuzzy match score} (FMS) function, which is usually based on the word-based edit distance~\citep{Wagner:1974:SCP:321796.321811}, and can be interpreted as the proportion of words common to $s$ and $s'$. The complement of the FMS is also used as an estimation of the post-editing effort, i.e. the proportion of word-level edit operations to perform on $t$ to get $t'$.

Even though deep learning methods have been recently applied to TMs, they have only been used to bias neural MT systems to produce translations similar to those in a TM \cite{tezcan2021towards}, or to devise alternative ways of computing the FMS between two source sentences \cite{ranasinghe-etal-2020-intelligent}. To the best of our knowledge, no approach has tried to overcome the main limitation of TMs: the reduced availability, as compared to target-language  (TL) monolingual corpora, of in-domain TMs for a given translation task. 


In this paper, we present an approach aimed at overcoming the aforementioned limitation by enabling the exploitation of in-domain TL monolingual corpora in CAT tools and increase the chance of getting \emph{useful} translation proposals. In such scenario, FMS cannot be computed because the TL sentences in the monolingual corpus have no source-language counterpart. We circumvent this issue by using cross-lingual sentence embeddings \cite{artetxe2019massively,reimers2020making,feng2020language} to project $t$ and $s'$ to the same space and comparing them by means of the cosine similarity. The resulting score, hereinafter \emph{neuroMatch}, allows us to obtain, for each $s'$ to be translated, a set of translation proposals from either the TM or the TL monolingual corpus. We then compute a form of FMS, which we term as \emph{neuroFMS}, with a COMET-based \cite{rei-etal-2020-comet} model, so that it can be used to estimate the  post-editing effort required to transform each translation proposal into  $t'$, the translation of $s'$. This allows professional translators to set a threshold for discarding proposals that would require too much post-editing effort, as they would do with FMS.

We have conducted a thorough experimentation with four language pairs, in which we study the utility of translation proposals retrieved with the conventional FMS function and with our neuroMatch alternative. The results show that neuroMatch successfully exploits TL monolingual corpora, similarly to the way FMS exploits TMs, and that it increases the amount of useful translation proposals in around 10 percentage points. Our evaluation also confirms that our neuroFMS alternative has a performance comparable to the conventional FMS, paving the way for an improved user experience, and enabling the combination and ranking of translation proposals obtained with both conventional FMS and neuroMatch. Finally, a human evaluation on one of the language pairs, in which professional translators were asked to assess the usefulness of the translation proposals obtained with FMS and neuroMatch, confirms the results of the automatic evaluation. This human evaluation also seems to indicate that the translation proposals retrieved with our approach are more useful than what the automatic evaluation shows.



The rest of the paper is organized as follows. Next section introduces the neuroMatch method for retrieving translation proposals from TL monolingual corpora, and the neuroFMS method for estimating a form of FMS. Section~\ref{s:experiments} then describes the experimental settings, whereas sections~\ref{s:results} and \ref{s:humaneval} present and discuss, respectively, the results of the automatic evaluation and of the human evaluation. Section~\ref{s:related_work} briefly describes the most relevant works in the literature, and Section~\ref{s:conclusions} summarizes the main conclusions. Finally, two additional sections discuss the limitations and ethical aspects of this work.

\section{Methodology}
\label{sec-methodology}

Our approach relies on two components: a method to efficiently retrieve translation proposals from a TL monolingual corpus (neuroMatch; Sect. \ref{se:retrieve-trans-proposal}), and a method, more expensive from the computational point of view, to estimate a form of FMS that can be used to estimate the post-editing effort of translation proposals  (neuroFMS; Sect.~\ref{se:estim-effort}).


\subsection{Retrieving translation proposals from target-language monolingual corpora}\label{se:retrieve-trans-proposal}
NeuroMatch builds on the use of cross-lingual sentence embeddings. There are a number of models to produce such embeddings in recent scientific literature~\citep{artetxe2019massively, reimers2020making, feng2020language}. In all cases, these models are trained to produce representations of sentences in a common $n$-dimensional space in which two similar sentences are placed close to each other, independently of the language in which they are written. Such feature enables the efficient search of candidate translations of a source sentence $s'$ in TL monolingual corpora.

In this work, we opted for LaBSE~\citep{feng2020language}, given that its embeddings are superior to other approaches \citep{reimers2020making} in the task of identifying sentence pairs which are mutual translations.
LaBSE is a neural architecture that consists of two paired BERT \cite{devlin-etal-2019-bert} encoders with shared weights that are pre-trained to perform both monolingual (masked language model) and bilingual (translation language model) self-supervised mask filling \citep{lample2019cross}. After training, in order to compute the similarity of two sentences (each one in a different language), each sentence is fed to one of the BERT encoders and the last-layer representations of the \texttt{[CLS]} token are selected as the corresponding sentence embeddings. Cosine similarity can then be used to compare them. In this way, the similarity measure integrates semantic knowledge, unlike the conventional, edit-distance-based FMS.

\subsection{Estimating fuzzy-matching scores}\label{se:estim-effort}

One of the key aspects behind the success of TM-based CAT tools is the twofold use of the conventional, edit-distance-based FMS: on the one hand, it is used to rank different translation proposals when more than one is available; on the other hand, its complement provides an estimation of the post-editing effort needed to convert the translation proposal $t$ into $t'$, an adequate translation of $s'$. This FMS is computed by comparing the sentence to be translated $s'$ and the source side sentence $s$ in the TU $(s,t)$; therefore, it cannot be computed for the translation proposals retrieved by neuroMatch from a TL monolingual corpus.

The cosine similarity used by neuroMatch can be effectively used to rank translation proposals, but cannot be used by translators to estimate the post-editing effort as they do with FMS: the cosine is bounded between -1 and 1 and has a Pearson's correlation coefficient with FMS around 0.8 (see Table \ref{tab:results-correl-fms}). To circumvent this problem and to allow for the seamless integration of translation proposals obtained from TMs and from monolingual corpora in the same CAT tool, we use neuroFMS, a COMET-based~\citep{rei-etal-2020-comet} model ---a neural model that builds on XLM-RoBERTa~\cite{conneau2020unsupervised}--- for the estimation of the fuzzy-matching score between $s'$ and the non-existent source-language sentence $s$ hypothetically paired with the TL sentence $t$ retrieved from the monolingual corpus. COMET has been successfully used both to evaluate and to estimate the quality of MT systems~\citep{rei-etal-2020-comet}; however, to the best of our knowledge, it has never been used in the context of TM-based CAT.


NeuroFMS is trained on a data set in which each instance consists of a source sentence $s'$, a translation proposal $t$, and the FMS between $s'$ and the source sentence $s$ paired with $t$; see next section for the details of how this data set is built. This model obtains the embeddings for $s'$ and $t$, and then uses them as the input to a feed-forward neural network that produces a prediction of the FMS.\footnote{Pre-trained COMET models are available, but they are conceived for a very different task ---prediction of direct quality assessment for MT---, hence the need for training an \emph{ad-hoc} model.}


It is worth noting that neuroFMS cannot be used to efficiently retrieve translation proposals from a monolingual corpus, as it would require scoring with COMET, for each source sentence to be translated $s'$, all pairs $\{(s',t_i)\}_{i=1}^N$, where $N$ is the number of sentences in the TL monolingual corpus. However, it may be a good option to estimate the post-editing effort of translation proposals retrieved with neuroMatch. NeuroMatch does not suffer from this problem because the embedding for each TL sentence can be pre-computed and cosine similarities can be efficiently computed by means of the search implementation over dense vectors provided by Faiss \citep{faiss}.

\section{Experimental setting} \label{s:experiments}
We evaluated our approach on four different language pairs, namely, English--Spanish (EN--ES), English--German (EN--DE), English--Czech (EN--CS) and English--Finnish (EN--FI), by simulating the translation of the source sentences in a test set when using a TM and, when applicable, a TL monolingual corpus. 

\paragraph{Data sets.} We used two sources of data for our experiments:  the DGT TM,\footnote{\url{https://ec.europa.eu/jrc/en/language-technologies/dgt-translation-memory}} a multilingual TM containing summaries of EU legislation, and the EurLex corpus,\footnote{\url{https://www.sketchengine.eu/eurlex-corpus}} a corpus containing all the legal documentation published in EurLex, the official journal of the European Commission and one of the sources from which the DGT-TM was built~\citep{pilos2014}. DGT-TM was used for building the test set and the TM to be used, whereas  EurLex was used as TL monolingual corpus. Both the DGT-TM and the EurLex corpus are provided at the document level. In both cases, the CELEX number of each document is provided; in this way, it is possible to warranty that no overlap exists between the test set and the TM or the monolingual corpus used.

Two different editions of the DGT-TM were used: DGT-TM 2020 was used as test set, using the source sentences as the sentences to be translated, and the target sentences as the reference translations for evaluation;  DGT-TM 2019 was used as the TM to query when translating the source sentences in the test set. 

Test sets were deduplicated, and those sentence pairs containing exactly the same text in the source and in the target languages were discarded, as in a real scenario they would remain untranslated. Source sentences containing only one word were also discarded, as they are likely to correspond to segmentation errors. Finally, sentence pairs in the test set with a source-target or target-source token ratio lower than 1/5 were  discarded, assuming that they are likely to correspond to alignment errors. Table \ref{tb:corpus} reports some statistics of the corpora used in our experiments.

\begin{table}[tb]
\centering
\def\arraystretch{0.95}
\setlength{\tabcolsep}{0.24em}
\begin{tabular}{c|l|r|r|r}

\textbf{Lang.} & \textbf{Resource} & \textbf{\# sents.} & \textbf{\# source} & \# \textbf{target}\\
\textbf{pair} &  &  & \textbf{tokens} & \textbf{tokens}\\
\hline \hline
\multirow{3}{*}{EN--ES} & test set & 174k & 3,831k & 4,437k \\
                       & TM & 430k & 7,490k & 8,619k \\
                       & mono & 56,476k & --- & 708,158k \\
\hline
\multirow{3}{*}{EN--DE} & test set & 167k & 3,666k & 3,341k \\
                       & TM & 425k & 7,478k & 6,741k \\
                       & mono & 60,102k & --- & 604,865k \\
\hline
\multirow{3}{*}{EN--CS} & test set & 184k & 3,869k & 3,253k \\
                       & TM & 428k & 7,467k & 6,170k \\
                       & mono & 46,811k & --- & 408,608k \\
\hline
\multirow{3}{*}{EN--FI} & test set & 180k & 3,752k & 2,716k \\
                       & TM & 433k & 7,489k & 5,346k\\
                       & mono & 53,619k & --- & 444,706k \\
\hline
\end{tabular}
\caption{For each language pair, amount of sentence pairs, source tokens and target tokens in the test set, in the TM and in the TL monolingual corpus used (mono).}
\label{tb:corpus}
\end{table}

\paragraph{Cross-lingual sentence embeddings.}
We used the implementation of LaBSE by \citet{feng2020language},\footnote{\url{https://tfhub.dev/google/LaBSE/1}} which provides language-agnostic cross-lingual sentence embeddings for 109 languages, and Faiss \cite{faiss}, which allows 
to efficiently compute cosine similarities in large sets of 
embeddings. 
These sentence  embeddings were used for computing similarity scores between the sentence to be translated $s'$ and the target sentences in the TM and in the TL monolingual corpus. 

\paragraph{Fuzzy-match score estimation.}

We explored two options: training an in-domain COMET model for our TM, and training an out-of-domain COMET model that could be used with any TM. While the first option is expected to perform better, it has the disadvantage of requiring the training of a new model for every new TM to be used. 

For the in-domain model, we built a training set through leaving-one-out  on the TMs used for the experiments, in such a way that, for every TU in the TM, we gathered an alternative translation proposal using the conventional FMS. The resulting training sets for each language pair were then concatenated to build a single multilingual training set. For training the out-of-domain model, we used the AutoDesk post-editing dataset.\footnote{\url{http://www.islrn.org/resources/290-859-676-529-5/}}  This dataset contains a collection of sentences from the IT domain ---clearly distant from the legal domain of our datasets--- and the best translation proposals retrieved from the TMs of the company, together with the corresponding FMS. Data is available for three out of the four language pairs in our experiments: EN--ES, EN--DE, and EN--CS. We concatenated them to train our COMET model. Although no EN--FI data is available, we expect a reasonable performance also for this language pair because XLM-RoBERTa, the underlying system in the COMET architecture, was trained, among others, on Finnish data.

For training the COMET models, we used the standard configuration as provided in the example files included in the COMET repository.\footnote{\url{https://github.com/Unbabel/COMET}} 

\paragraph{Evaluation.}



According to \citet{bowker2002computer,mitkov-2015}, the users of TM-based CAT tools usually set a threshold for the FMS above 60\% so that the CAT tool does not bother them with TUs with low FMS: in general, translators consider translation proposals \emph{useful} if they have to post-edit less than 40\% of the sentence. Following this idea, we computed the translation edit rate (TER; \citet{snover2006study}) between every translation proposal $t$ and its corresponding reference translation, $t'$, and considered useful those proposals with a TER below 0.4. TER measures the proportion of edits operations 
needed to convert $t$ into~$t'$.

It is worth noting that 
computing TER for the whole test set and for all the translation proposals retrieved, regardless of their quality, would have reported inconclusive results because of the noise introduced by the low-quality  translation proposals. 

\section{Results and discussion} \label{s:results}
First, we discuss the results obtained when comparing the conventional, edit-distance based FMS and the LaBSE-based neuroMatch (Section \ref{s:result-labse}). After that, we evaluate the FMS estimation based on COMET (neuroFMS; Section \ref{s:results-comet}).


\subsection{Neural fuzzy matching}\label{s:result-labse}

\begin{table*}[tb]
\centering
\begin{tabular}{l|l|r|r|r|r}
 \multicolumn{1}{c|}{\textbf{Approach}} & \multicolumn{1}{c|}{\textbf{Resources}} &
 \multicolumn{1}{c|}{\textbf{EN--ES}} & \multicolumn{1}{c|}{\textbf{EN--DE}} & \multicolumn{1}{c|}{\textbf{EN--CS}} & \multicolumn{1}{c}{\textbf{EN--FI}} \\
\hline \hline
FMS                    & TM       & 21.3\%  & 20.3\%  & 18.7\% & 20.6\%  \\
neuroMatch & TM       & 20.2\%  & 18.8\%  & 17.7\% & 18.8\%  \\ \hline
neuroMatch & mono     & 28.8\%  & 24.2\%  & 21.8\% & 24.6\%  \\
neuroMatch & TM+mono  & \textbf{34.7\%}  & \textbf{30.9\%}  & \textbf{27.4\%} & \textbf{30.1\%}  \\
\hline 
\end{tabular}
\caption{Percentage of sentences in the test set for which a useful translation proposal is obtained, i.e., a proposal with a TER below 0.4. The second column indicates the resources being exploited: TM, TL monolingual corpora (mono) or both (TM+mono).} 
\label{tb:allsystems}
\end{table*}

Table~\ref{tb:allsystems} reports the percentage of sentences in the test set for which a useful proposal (see above) is found when using FMS and neuroMatch to obtain translation proposals. 
We have used neuroMatch for retrieving translation proposals from the very same TM used with FMS, as well as for retrieving translation proposals from the TL monolingual corpus (mono) and from both (TM+mono). In all cases, neuroMatch computes the similarity between $s'$ and a target sentence (either from the TM or from the TL monolingual corpus).

When compared with FMS, neuroMatch is able to retrieve a similar amount of useful translation proposals from the TM (1.4 percentage points less on average), and a larger amount (4.6 percentage points more on average) when exploiting just the TL monolingual corpus. Exploiting both the TM and the TL monolingual corpora, neuroMatch is able to retrieve a useful translation proposal for around one third of the sentences in the test set (on average, 10.6 percentage points more than FMS). Averaging over all language pairs, 19\% of these translation proposals come from the TM, and 81\% from the TL monolingual corpus.

A comparison of FMS and neuroMatch when both exploit just the TM shows that FMS provides a slightly higher number of useful translation proposals than neuroMatch. However, there is a relevant detail that is worth noting: for around 20\% of the instances of the test set for which both FMS and neuroMatch provide a useful translation proposal, the one obtained with neuroMatch is better (has a lower TER) than that obtained with FMS.\footnote{The exact percentage per language pair are as follows: 26\% for EN--ES, 24\% for EN--DE, 17\% for EN--CS and 20\% for EN--FI.} This means that both FMS and neuroMatch could be used together in a CAT environment
to improve the way in which TMs are used for translation. 

\begin{figure*}[tb]
  \centering
  \includegraphics[width=0.49\linewidth]{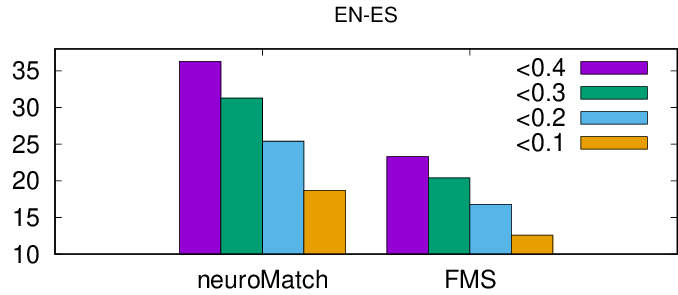}
  \includegraphics[width=0.49\linewidth]{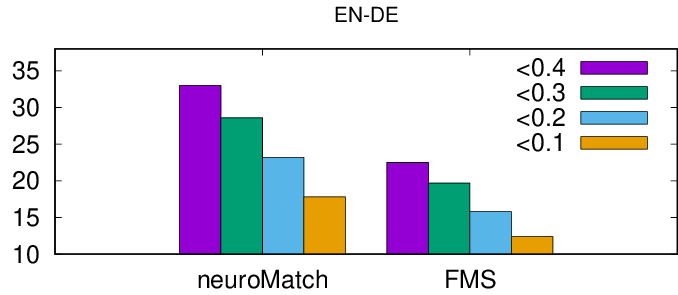}
  \includegraphics[width=0.49\linewidth]{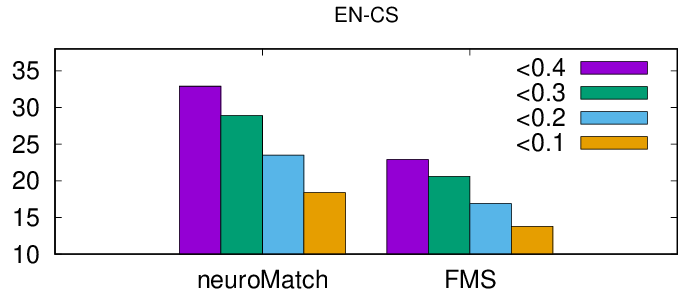}
  \includegraphics[width=0.49\linewidth]{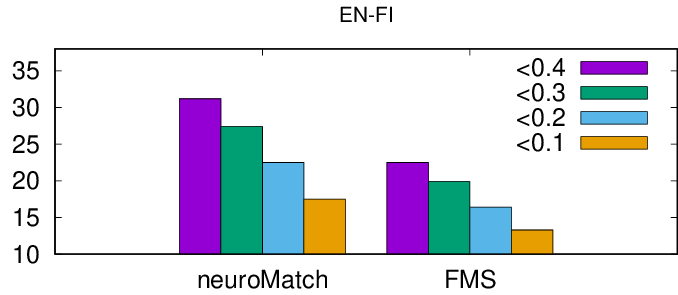}
\caption{Percentage of sentences in the test set for which a translation proposal is obtained with FMS on TM and with neuroMatch on TM+mono when TER is below 0.4, 0.3, 0.2 and 0.1.}
  \label{fig:winningApp}
\end{figure*}

Figure~\ref{fig:winningApp} provides a more detailed analysis of the translation proposals retrieved with FSM and neuroMatch when the latter searches for translation proposals both in the TM and in the TL monolingual corpus (TM+mono). The figure shows the percentage of translation proposals provided by each
method for each language pair and for different thresholds of TER,
in particular 0.4, 0.3, 0.2 and 0.1. Recall that TER is computed by comparing the translation proposal and the reference translation in the test set. As can be seen, neuroMatch outperforms FMS in all cases.

\subsection{Fuzzy-match score estimation}\label{s:results-comet}

Table~\ref{tab:results-correl-fms} shows the Kendall rank correlation coefficient ($\tau$) and the Pearson's correlation coefficient ($\rho$) between FMS and neuroMatch (the cosine similarity between LaBSE embeddings; first row of the table), and between FMS and neuroFMS, both in-domain and out-of-domain COMET models (second and third rows). These results were computed on the translation proposals obtained from the TM, as the FMS cannot be computed for translation proposals retrieved from the TL monolingual corpus.

\begin{table}[tb]
\centering
\setlength{\tabcolsep}{0.19em}
\begin{tabular}{@{}l|rr|rr|rr|rr}
  & \multicolumn{2}{c|}{\textbf{EN--ES}} & \multicolumn{2}{c|}{\textbf{EN--DE}} & \multicolumn{2}{c|}{\textbf{EN--CS}} & \multicolumn{2}{c}{\textbf{EN--FI}} \\
  & \multicolumn{1}{c}{$\tau$} & \multicolumn{1}{c|}{$\rho$} & \multicolumn{1}{c}{$\tau$} & \multicolumn{1}{c|}{$\rho$} & \multicolumn{1}{c}{$\tau$} & \multicolumn{1}{c|}{$\rho$} & \multicolumn{1}{c}{$\tau$} & \multicolumn{1}{c}{$\rho$} \\ \hline \hline
\textbf{neuroMatch}                 & .55 & .83 & .54 & .81 & .58 & .81 & .55 & .83 \\
\textbf{neuroFMS out}            & .72 & .89 & .70 & .88 & .68 & .88 & .70 & .89 \\ 
\textbf{neuroFMS in}     & \textbf{.78}  & \textbf{.93}  & \textbf{.77}  & \textbf{.92} & \textbf{.73} & \textbf{.91} & \textbf{.77} & \textbf{.92} \\
\hline 
\end{tabular}
\caption{For each language pair, Kendall's ($\tau$) and Pearson's correlation ($\rho$) coefficients between the FMS and neuroMatch (first row), and the neuroFMS COMET-based models trained on in-domain (neuroFMS in) and on out-of-domain (neuroFMS out) data.} 
\label{tab:results-correl-fms}
\end{table}

As can be seen, neuroFMS highly correlates with FMS and, as expected, the in-domain model performs better than the out of domain model, but the results are close enough to consider using the out-of-domain model because it avoids training a new model for every new domain; recall that the same model is used for all language pairs. The table also shows that neuroMatch correlates worse with FMS than neuroFMS; although it also shows a high correlation. It is worth noting that, even if neuroMatch highly correlates with FMS, it cannot be directly used by translators to estimate post-editing effort because its values are in the interval $[-1,1]$ rather than in $[0,1]$ as it is the case of FMS and neuroFMS. 
    
To further prove the usefulness of neuroFMS, we decided to run an extrinsic evaluation in which the neuroFMS score was used to combine translation proposals obtained with FMS and neuroMatch (TM+mono). As already mentioned, in some cases, both FMS and neuroMatch can provide useful translation proposals, but one may be better than the other. We took the subset of instances from the test set for which either FMS or neuroMatch (or both) provided a useful translation proposal and computed the total TER when using FMS, neuroMatch, and the combination of both using neuroFMS to decide which translation proposal is better. Results are shown in Table~\ref{tab:combining-methods}. It also includes an oracle row as an upper-bound reference. The results obtained clearly show that the neuroFMS models are able to improve the result of FMS and neuroMatch in isolation, as using them to combine the translation proposals provided by the two methods lead to better results, especially for the in-domain model. In any case, there is room for improvement, as the best results are still far from the oracle.


\begin{table}[tb]
\centering
\setlength{\tabcolsep}{0.17em}
\begin{tabular}{l|c|c|c|c}
   & \textbf{EN--ES} & \textbf{EN--DE} & \textbf{EN--CS} & \textbf{EN--FI} \\ \hline \hline
   \textbf{FMS} & .44 & .44 & .41 & .42 \\
   \textbf{neuroMatch} & .18 & .19 & .17 & .19 \\
   \textbf{neuroFMS out} & .16 & .17 & .15 & .17 \\
   \textbf{neuroFMS in} & \textbf{.16} & \textbf{.16} & \textbf{.15} & \textbf{.15} \\
   \rowcolor{Gray}
   \textbf{Oracle}  & .13  & .13 & .11 & .11 \\
  \hline
\end{tabular}
\caption{TER for the instances in the test set for which a useful translation proposal is found with FMS, neuroMatch or both. NeuroMatch was used to obtained translation proposal from the TM and the TL monolingual corpus (TM+mono). When more than one proposal is found, neuroFMS is used to choose the best one. Last row shows the results provided by an oracle using the reference translation (upper bound).}
\label{tab:combining-methods}
\end{table}

\section{Human evaluation} \label{s:humaneval}
To complement the results obtained with the automatic evaluation, an experiment was conducted with professional translators and a random sample of the EN--ES data used for the automatic evaluation.
%
Six professional translators were provided with a form containing a small collection of English sentences and two translation proposals in Spanish: one retrieved from the TM using FMS, and another retrieved from the TM plus the TL monolingual corpus using neuroMatch. Translators did not know the method used to obtain each translation proposal. They were asked to indicate 
which translation proposals they would use as a draft translation to be post-edited, or if they would rather translate the source sentence from scratch. In those cases in which they considered both translation proposals acceptable for post-editing, they could indicate if one of them would require less post-editing effort (they could also indicate that both were of similar quality).

The data sets given to each translator contained about 350 instances, and were distributed in such a way that half of the instances provided to each translator were also annotated by other translators, the other half were not. In this way, we were able to  compute the inter-annotator agreement between every pair of translators (see Table~\ref{tab:interanotator-areement}). In addition, 15 control instances were provided to each translator to make sure that they understood the task and that their annotations are valid. Adding up all the data instances provided to the six translators, the data set contained 2,130 instances (from which 1,035 were unique). 

Regarding the interanotator agreement, 
Table~\ref{tab:interanotator-areement} 
shows a high agreement between annotators \#0, \#2, \#3 and \#4; annotators \#1 and \#5 present a slightly lower value of the Cohen's kappa. In any case, all the values range from moderate to substantial agreement according to \citet{landis1977measurement}.   

\begin{table}[tb]
\centering
\setlength{\tabcolsep}{0.5em}
\begin{tabular}{l|ccccc}

   & \textbf{An. 1} & \textbf{An. 2} & \textbf{An. 3} & \textbf{An. 4} & \textbf{An. 5}  \\ \hline \hline
   \textbf{An. 0}          & .53 & .71 & .77 & .73 & .58 \\
   \textbf{An. 1} & --- & .47 & .44 & .69 & .60 \\
   \textbf{An. 2} & --- & --- & .68 & .74 & .60 \\
   \textbf{An. 3} & --- & --- & --- & .65 & .57 \\
   \textbf{An. 4}  & --- & --- & --- & --- & .68 \\
  \hline 

\end{tabular}
\caption{Inter-annotator agreement (Cohen's Kappa) between the six translators that participated in the human evaluation (see running text).}
\label{tab:interanotator-areement}
\end{table}

After analysing the assessments provided by the translators, neuroMatch was considered the best choice for 856 instances (~40\% of the instances in the data set), while FMS was the best choice for 388 instances (~18\% of the instances). Both were considered equally useful for 132 instances (~6\% of the instances). These figures confirm the results of the automatic evaluation and seem to indicate that neuroMatch could be more useful than what the automatic evaluation showed.


These human annotations also allowed us to  evaluate the scores obtained with neuroFMS when combining FMS and neuroMatch. We took the subset of instances for which one of the systems was considered better than the other (1,244 instances in total) and checked if neuroFMS and the professional translators agreed. In about 80\% of the cases, neuroFMS  and the human annotators agreed; this result is also in line with the automatic evaluation reported in previous section.



\section{Related work}
\label{s:related_work}


In this section, we review the related work as regards the computation of sentence embeddings (Section~\ref{sect-sentence-embedding}), and the exploitation of these embeddings in the context of TMs (Section~\ref{sect-embeddins-tm}).

\subsection{Sentence embeddings}
\label{sect-sentence-embedding}

The use of word embeddings, that is, vector representations of words showing mathematical properties that correlate with semantic similarities, can be traced back to many decades ago~\citep{lsa1990}. The 2010s saw the emergence of a number of self-supervised techniques that allowed for more powerful word representations, first in the form of non-contextual vectors~\citep{mikolov2013distributed, pennington-etal-2014-glove, bojanowski2017enriching}, and afterwards as context-sensitive representations obtained by processing sentences through recurrent~\citep{peters2018deep} or transformer-based~\citep{lample2019cross,liu2019roberta} neural architectures, being BERT~\citep{devlin-etal-2019-bert} possibly the most well-known and exploited example of the attention-based approaches.

Sentence embeddings may be seen as an extension of the main ideas behind word embeddings to the case of variable-length inputs. They were first computed~\cite{le2014distributed} by extending the method used to obtain non-contextual word embeddings, but soon they evolved to more elaborated architectures such as the encoder-decoder system exploited by skip-thought vectors~\cite{skipthoughts2015} to encode a representation of input sentences that allowed the decoder to generate the preceding and following sentences. Current methods mostly use attention-based encoders to attain more elaborated representations. Some models add a classifier performing a natural language inference task (for example, detecting contradiction, entailment, or neutrality in sentence pairs) that gets fed with pooled representations of the encoder embeddings, the whole system trained in an end-to-end fashion. Other models directly train the encoder on some self-supervised task. As regards models explicitly addressing semantic tasks on sentence pairs, \citet{conneau2017supervised} considered two separate encoders, but~\citet{ranasinghe-etal-2019-semantic} suggested using a dual-encoder framework. \citet{reimers2019sentence} followed a similar semantic-task-mediated setting to introduce Sentence-BERT, where the dual encoder consisted of two paired BERT models with shared weights. State-of-the-art systems include SimCSE~\cite{simcse} which exploits contrastive learning to move the embeddings of similar sentences closer, and Sentence-T5~\cite{st5} that obtains best results with a full encoder-decoder architecture.

By integrating data in different languages, some of the previous models can be extended to generate cross-lingual sentence embeddings, either as simple word-level embeddings~\citep{conneau2017word} or as general-purpose sentence-level embeddings such as those provided by mBERT~\citep{pires2019multilingual}, LASER~\citep{artetxe2019massively}, m-USE~\citep{yang2020multilingual} or XML-R~\citep{conneau2020unsupervised}, among others. Our approach uses LaBSE (which stands for language-agnostic BERT sentence embeddings) whose embeddings are optimised~\citep{feng2020language} to be similar for bilingual sentence pairs that are mutual translation, a result of the fact that the encoder has been pre-trained with the \emph{translation language model} self-supervised objective which performs mask filling in the output given as input the concatenation of two parallel sentences in two languages.



\subsection{Integration of embeddings in translation memories}
\label{sect-embeddins-tm}

Most of the works involving neural-based sentence similarities and translation aim at improving neural MT, not TMs. Given a new sentence $s'$ to be automatically translated first and post-edited next, \citet{farajian2017multi} propose finding its closest sentence $s$ in a TM via a non-neural similarity measure, and then using the TU $(s,t)$ to fine-tune an existing neural MT model before using it to provide the translator with a draft translation. Similarly, \citet{gu2018search} use FMS on the source side to retrieve sentence pairs from the TM; neural embeddings are then obtained for each of these pairs and stored in an external memory which is coupled to a neural MT system that translates $s'$ into the target language. \citet{zhang2018guiding} promote in the decoder of an neural MT system outputs that contain $n$-grams appearing in translations retrieved from the TM. \citet{bulte-tezcan-2019-neural} introduced neural fuzzy-match repair in which an neural MT system that can have the best translation proposals in the TM optionally appended to its input. \citet{cao2018encoding} also add TL sentences to \emph{inspire} the neural MT system, but source-language inputs and TL inputs are processed by different encoders and the resulting representations combined by a gating mechanism. Finally, the approach by~\citet{cai2021neural} shares some elements with ours, but again their objective is to improve neural MT: they also use a dual-encoder to learn cross-lingual embeddings; the resulting representations for relevant translation proposals in the TM are then stored in a memory which is connected to the transformer via attention mechanisms in order to guide the translation process. 

To our knowledge, the most similar approach to ours in the literature is the one proposed by~\citet{ranasinghe-etal-2020-intelligent}: they use neural-based sentence similarities to retrieve from the TM the TUs $(s,t)$ whose source sentences $s$ are close to the sentence to be translated $s'$. However, they perform the search in the source-language space which prevents the integration of additional monolingual text. In spite of this critical difference they also find neural translation proposals to be better than FMS translation proposals for values of the FMS below 80\%.

\section{Conclusions} \label{s:conclusions}
We have presented a novel neural approach to overcome the main limitation of TM-based CAT tools: the reduced availability, as compared to TL monolingual corpora, of in-domain TMs. Our approach consists of two different neural components: neuroMatch and neuroFMS. NeuroMatch uses cross-lingual sentence embeddings to search for translation proposals in TL monolingual corpora. NeuroFSM uses an \emph{ad-hoc} COMET model to estimate a form of FMS between the source sentence to be translated and the translation proposals retrieved with neuroMatch. 

We have extensively evaluated our approach on four different language pairs using an automatic evaluation metric, and found out that our approach is able to retrieve 10 percentage points more \emph{useful} translation proposals than the conventional FMS. In line with common practice by professional translators, we considered useful those translation proposals with a translation edit rate (TER) below 0.4.
We also studied the performance of our approach with more restrictive definitions of usefulness, i.e. with TERs of 0.3, 0.2 and 0.1,
and in all cases neuroMatch retrieves more useful translation proposals than FMS. A human evaluation on one language pair confirms these results.

As regards neuroFMS, it has shown a high correlation with FMS and that when it is used to combine translation proposal obtained with FMS and neuroMatch, the TER of the set of sentences for which a translation proposal is found with either method is reduced. This last result opens the door to further improvements in professional translation productivity. Both conventional FMS and neuroMatch can be seamlessly integrated: both types of translation proposals can be ranked together with neuroFMS, and professional translators can then use neuroFMS to estimate the postediting effort of the different translation proposals they are offered, as they do with FMS. 



\section*{Limitations}  \label{s:limitations}
Our experimentation has some limitations that are worth to be taken into account for future work:

\paragraph{Computing TER regarding an independent reference translation.} In this paper, the TER between a translation proposal and an independent reference is used as a metric of the usefulness of the translation proposal (in most experiments, proposals with a TER lower than 0.4 are considered useful). However, a valid translation proposal could be discarded if the reference to which it is compared to is formulated in a very different way (order of words, use of synonyms, etc.). This fact could lead us to underestimate the quality of the translation proposals evaluated. A much better option would be to use post-edited versions of the translation proposals (human TER, or HTER); unfortunately, producing such data set would be prohibitively expensive for the size of the test sets used. In any case, using independent references still give us a reasonable estimation of the performance and, somehow, the results obtained are a lower bound of the actual performance of the methods evaluated.


\paragraph{Human evaluation only on a language pair.} While automatic evaluation was conducted on four language pairs, human evaluation, which was substantially more expensive, was only carried out for one of them. While we could not afford the cost of a larger human evaluation, we consider that it is reasonable to assume that the conclusions obtained for this language pair can be extrapolated to the rest of language pairs used in the automatic evaluation.

\paragraph{Application of the methods described to other languages.} It is worth noting that the models used in this paper are pre-trained for a high number of languages: 109 in the case of LaBSE, and 100 languages for XLM-RoBERTa (the encoder used for COMET models). However, the truth is that there are languages that are not covered by these models. On the other hand, the general ideas included in this work could be applied to languages not covered by these models by replacing them with other similar models.

\paragraph{Further combination of neuroMatch and neuroFMS.} We tried to keep the evaluation of neuroMatch and neuroFMS as separate as possible in this work, as we consider that this allows to better understand the performance and contribution of each model to the proposed CAT scenario. However, it would be possible to combine them in smarter ways. For example, it would be possible to get the $n$-best translation proposals using neuroMatch and then use neuroFMS to keep only the best one. Such evaluation will be covered in future works. 

\section*{Acknowledgements}
This paper is part of the R+D+i project PID2021-127999NB-I00 funded by the Spanish Ministry of Science and Innovation (MCIN), the Spanish Research Agency (AEI/10.13039/501100011033), and the European Regional Development Fund A way to make Europe. The computational resources used for the experiments were funded by the European Regional Development Fund through project IDIFEDER/2020/003. We thank the European Association for Machine Translation for funding the human evaluation reported in this paper through the EAMT Sponsorship of Activities for 2021.


\section*{Ethical considerations}
All the experiments reported in this paper build on existing datasets that are publicly available. The software used to run the experiments reported is available in the following repository: \url{https://github.com/transducens/CrossLingualNeuralFMS}. Appendix~\ref{a:repro} provide details regarding the reproducibility of the experiments.

The experiment described in Section~\ref{s:humaneval} involve professional translators. They were hired according to the Spanish labour legislation and were paid a fair salary for their work. Before starting the experiment, they were properly informed about the purpose of the experiment and the fact that the results of their work would be published.

\appendix

\section{Reproducibility} \label{a:repro}
The code that allows to download the data and to run the experiments presented in Section~\ref{s:experiments} is available in the repository \url{https://github.com/transducens/CrossLingualNeuralFMS}. The only exception to this is the EurLex monolingual corpus, which was downloaded from  \url{https://www.sketchengine.eu/eurlex-corpus/}. This corpus is provided by Sketch Engine under request with Creative Commons BY-NC-SA 2.0 license.\footnote{\url{https://creativecommons.org/licenses/by-nc-sa/2.0/}} 

We also provide the configuration file used to train the COMET models evaluated in Section~\ref{s:results-comet}. A general description of this model is provided in Appendix~\ref{a:comet}.

The results of the human evaluation are included in the repository mentioned above. They are provided in a tab-sepparated-value format, including the source sentences, translation proposals evaluated, and assessment provided by the translators.

\section{COMET model used for neuroFMS} \label{a:comet}
The \emph{ad-hoc} COMET model used for estimating fuzzy-matching scores (see  Section~\ref{s:experiments}) was trained using the referenceless-model configuration file from the COMET repository,\footnote{\url{https://github.com/Unbabel/COMET}} commit \texttt{c9e1818583c1677328e3d1b649bb6c9c6}. The most relevant parameters are:
\begin{itemize}
    \item XLM-RoBERTa large with frozen embeddings as base model;
    \item AdamW optimizer with learning rate of 3.1e-05;
    \item dropout of 0.15;
    \item batch size of 4 sentence pairs;
    \item a feedforward with two hidden layers (sizes 2048 and 1024) at the end; and
    \item sentence embeddings built by obtaining the average of all layers (pooling).
\end{itemize}
Only one parameter was modified in the standard configuration: a sigmoid activation function was used in the output layer so that the prediction is in (0,1).

\section{Computational resources}
To run the experiments reported in this paper, we used a computer with 8 GPUs Geforce RTX 2080 TI. All the experiments involving neuroFMS (that uses LaBSE to build embeddings and Faiss to search for translation proposals) substantially benefit from using this computational infrastructure. With this infrastructure, building the embeddings for the segments in the test set, the TM and the monolingual corpus took about 12 hours; searching for the best translation proposals with this technique took about 2 hours. Both LaBSE and Faiss can be run on CPU instead of GPU, but the time required is substantially higher.

Collecting translation proposals from the TM using FMS took about a week on a machine equipped with 96 Intel(R) Xeon(R) Gold 6252 CPUs at 2.10GHz.

Training our ad-hoc COMET models was not possible on the Geforce RTX 2080 TI GPUs mentioned above, as they do not have enough memory to load the models used by COMET. For this task, we used an Nvidia A100 GPU; training each model took about 20 hours.

\bibliography{custom}
\bibliographystyle{acl_natbib}

\end{document}